\documentclass[10pt,journal,compsoc]{IEEEtran}
\usepackage{times}
\usepackage{makecell}
\usepackage{epsfig}
\usepackage{graphicx}
\usepackage{amsmath}
\usepackage{amssymb}
\usepackage{rotating}
\usepackage{multirow}
\usepackage{hhline}

\def\ff{\mathbf{f}}

\def\ssc{\mathbf{s}}
\def\vv{\mathbf{v}}
\def\ee{\mathbf{e}}
\def\OO{\mathbf{1}}
\def\R{{\rm I\!R}}   % the real numbers
\DeclareMathOperator{\argmax}{max}
\newcommand*{\Scale}[2][4]{\scalebox{#1}{$#2$}}%
\ifCLASSOPTIONcompsoc
  \usepackage[nocompress]{cite}
\else
  \usepackage{cite}
\fi

\ifCLASSINFOpdf
\else
\fi

\begin{document}
\title{Deep Simplex Classifier for Maximizing the Margin in Both Euclidean and Angular Spaces}
\author{Hakan~Cevikalp,~Hasan~Saribas \\ 
\thanks{H. Cevikalp is with the Department
of Electrical and Electronics Engineering, Eskisehir Osmangazi University,
Eskisehir, Turkey e-mail: hakan.cevikalp@gmail.com.}
\thanks{H. Saribas is with Huawei Turkey R\&D Center, AIE Department, Istanbul, Turkey e-mail: hasan.saribas1@huawei.com.}	
}

\IEEEtitleabstractindextext{%
\begin{abstract}
The classification loss functions used in deep neural network classifiers can be grouped into two categories based on maximizing the margin in either Euclidean or angular spaces. Euclidean distances between sample vectors are used during classification for the methods maximizing the margin in Euclidean spaces whereas the Cosine similarity distance is used during the testing stage for the methods maximizing margin in the angular spaces. This paper introduces a novel classification loss that maximizes the margin in both the Euclidean and angular spaces at the same time. This way, the Euclidean and Cosine distances will produce similar and consistent results and complement each other, which will in turn improve the accuracies. The proposed loss function enforces the samples of classes to cluster around the centers that represent them. The centers approximating classes are chosen from the boundary of a hypersphere, and the pairwise distances between class centers are always equivalent. This restriction corresponds to choosing centers from the vertices of a regular simplex. There is not any hyperparameter that must be set by the user in the proposed loss function, therefore the use of the proposed method is extremely easy for classical classification problems. Moreover, since the class samples are compactly clustered around their corresponding means, the proposed classifier is also very suitable for open set recognition problems where test samples can come from the unknown classes that are not seen in the training phase. Experimental studies show that the proposed method achieves the state-of-the-art accuracies on open set recognition despite its simplicity.
\end{abstract}

}

\maketitle

\IEEEdisplaynontitleabstractindextext

% \fi
\IEEEpeerreviewmaketitle

\IEEEraisesectionheading{\section{Introduction}\label{sec:introduction}}

\IEEEPARstart{D}{eep} eural network classifiers have been dominating many fields including computer vision by achieving state-of-the-art accuracies in many tasks such as visual object, activity, face and scene classification. Therefore, new deep neural network architectures and different classification losses have been constantly developing. The softmax loss function is the most common function used for classification in deep neural network classifiers. Although the softmax loss yields satisfactory accuracies for general object classification problems, its performance for discrimination of the instances coming from the same class categories (e.g., face recognition) or open set recognition (a classification scenario that allows the test samples to come from the unknown classes) is not satisfactory. The performance decrease is typically attributed to two factors: there is no mechanism for enforcing large-margin between classes and the softmax does not attempt to minimize the within-class scatter which is crucial for the success in open set recognition problems. 

To improve the classification accuracies of the deep neural network classifiers, many researchers focused on maximizing the margin between classes. The recent methods can be roughly divided into two categories based on maximizing the margin in either Euclidean or angular spaces. The methods targeting margin maximization in the Euclidean spaces attempt to minimize the Euclidean distances among the samples coming from the same classes and maximize the distances among the samples coming from different classes. Euclidean distances are used during testing stage after the network is trained. In contrast, the methods that maximize the margin in the angular spaces use the cosine distances for classification. 

To maximize the margin in Euclidean space, \cite{R1,R2} combined the softmax loss function with the center loss for face recognition. Center loss reduces the within-class variations by minimizing the distances between the individual face class samples and their corresponding class centers. The resulting method significantly improves the accuracies over the method using softmax alone in the context of face recognition. 
A variant of the center loss called the contrastive center loss \cite{R3} minimizes the Euclidean distances between the samples and their corresponding class centers and maximizes the distances between samples and the centers of the rival (non-corresponding) classes. \cite{R10} combined the range loss with the softmax loss to maximize the margin in the Euclidean spaces. %The range loss consists of two terms that simultaneously minimize the within-class variations and maximize the between-class separation. The within-class variation is minimized through penalizing the maximum harmonic range of each class whereas the between-class separation is maximized by penalizing the rival class centers that have the minimum distance in a given batch. 
%Wei et al. \cite{R11} combined softmax loss  and center loss functions with the minimum margin loss where the minimum margin loss enforces all class center pairs to have a distance larger than a specified threshold.  
\cite{R12} introduced a method using softmax loss function with the marginal loss to create compact and well separated classes in Euclidean space. Marginal loss enforces the distances between sample pairs from different classes to be larger than a selected threshold while enforcing the distances between sample pairs coming from the same classes to be smaller than the selected threshold. 
\cite{R50} proposed a deep neural network based open set recognition method that returns compact class acceptance regions for each known class. %In this framework, hinge loss and polyhedral conic functions are used for the between-class separation. 
The methods using Contrastive loss minimize the Euclidean distances of the positive sample pairs and penalize the negative pairs that have a distance smaller than a given margin threshold. In a similar manner, \cite{R6,R7,R8,R9} employ triplet loss function for the same purpose.
%that used a positive sample, a negative sample and an anchor. An anchor is also a positive sample, thus the within-class compactness is achieved by minimizing the Euclidean distances between the anchor and positive samples whereas the distances between anchor and negative samples are maximized for between-class separation. 
Although methods using both contrastive and triplet loss functions return compact decision boundaries, they have limitations in the sense that the number of sample pairs or triplets grows quadratically (cubicly) compared to the total number of samples, which results in slow convergence and instability. A careful sampling/mining of data is required to avoid this problem. 
Overall, the majority of the methods maximizing margin in the Euclidean spaces have shortcomings in a way that they are too complex since the user has to set many weighting and margin parameters. %This is due to the fact that the main classification loss functions include many terms that needs to be properly weighted. Furthermore, many of these methods are not suitable for open set recognition problems since they do not return compact acceptance regions for classes. 

The methods that enlarge the margin in the angular spaces typically revise the classical softmax loss functions to maximize the angular margins between rival classes, and almost all methods are especially proposed for face recognition. 
To this end, \cite{R14,R15} proposed the SphereFace method which uses the angular softmax (A-softmax) loss that enables to learn angularly discriminative features. %The proposed method projects the original Euclidean space of features to an angular space, and introduces a multiplicative angular margin to maximize the between-class separation. %However, the SphereFace  method has limitations in the sense that the decision margins change for each class. As a result, some between-class features have a larger margin while others have a smaller margin, which reduces the discriminating power \cite{RR16}. 
\cite{R17} proposed the RegularFace method in which A-softmax term is combined with an exclusive regularization term to maximize the between-class separation. \cite{R16} introduced the CosFace method which imposes an additive angular margin on the learned features. To this end, they normalize both the features and the learned weight vectors to remove radial variations and then introduce an additive margin term, $m$, to maximize the decision margin in the angular space. 
A similar method called ArcFace is introduced in \cite{R18}, where an additive angular margin is added to the target angle to maximize the separation in angular space. %This method has been extended in \cite{RR28} such that each class is represented with several subclusters instead of a single one.
%Liu et al. \cite{RR22} proposed AdaptiveFace method that enables to adjust the margins for different classes adaptively.
%\cite{RR20} introduced uniform loss function to learn equidistributed representations for face recognition. %By using both A-softmax and uniform loss functions together, the class centers of the learned CNN features are uniformly spread on a hypersphere manifold.
We would like to point out that almost all methods that maximize the margin in the angular space are proposed for face recognition. As indicated in \cite{R50}, these methods work well for face recognition since face class samples in specific classes can be approximated by using linear/affine spaces, and the similarities can be measured well by using the angles between sample vectors in such cases. Linear subspace approximation will work as long as the number of the features is much larger than the number of class specific samples which holds for many face recognition problems. However, for many general classification problems, the training set size is much larger compared to the dimensionality of the learned features and therefore these methods cannot be generalized to the classification applications other than face recognition. In addition to this problem, these methods are also complex since they have many parameters that must be set by the user as in the methods that maximize the margin in the Euclidean spaces.

The methods that are most closer to the proposed methodology are proposed in \cite{RNN3,RNN1,RNN2}. These methods introduce loss functions for learning uniformly distributed representations on the hypersphere manifold through potential energy minimization. However, these studies consider the layer regularization problem rather than direct classification problem and apply hyperspherical uniformity to the learned weights. The main idea is to learn diverse deep neural network weights that are uniformly distributed on a hypersphere in order to reduce the redundancy. Therefore, these methods are more complex (in some sense it is also more sophisticated since it applies the hyperspherical uniformity to all neural network layers). Consequently, there are many hyperparameters that must be fixed in the resulting method. In contrast, our proposed method is simple and there is no hyperparameter to tune. Also, when this idea is used in the classification layer, the distances between the resulting class representative weights are not equivalent as in our proposed method. A related study called UniformFace \cite{RR20} used the same idea in the classification layer only and introduced uniform loss function to learn equidistributed representations for face recognition. Another similar method using class centroids is introduced in \cite{RNN4} for distance metric learning. Although this study focuses on distance metric learning, it uses class centers chosen as the basis vectors of $C$-dimensional space as anchors. Then, as in triplet loss, it attempts to minimize the distances between the data samples and the corresponding class centers and to maximize the distances between the samples and rival class centers. The selected class centers are fixed as in our proposed method and it has a restriction that the feature dimension size must be larger than or equal to the number of classes similar to our case. Compared to this method, our proposed method is much simpler and run-time complexity of the proposed method is significantly less. Moreover, the authors make 2 critical mistakes in their proposed method: The first mistake is to choose the centers from the surface of a unit hypersphere (a hypersphere with radius 1). As we discussed below, the data samples lie near the surface of a growing hypersphere as the dimension increases. Therefore, setting the hypersphere radius to 1 is wrong for high-dimensional feature spaces, and similar findings are reported in studies such as ArcFace \cite{R18} and CosFace \cite{R16}. The second mistake is to use a fully connected layer alone for increasing the dimensionality when the feature dimension is smaller than the number of classes. A fully connected layer just uses the linear combination of existing features and the resulting space has the same dimensionality as in the original feature space (this issue is explained in more details below). As a result, the dimensionality is not increased, and this method will not work for large-scale problems where the number of classes is very large. 

\smallskip\noindent{\bf Contributions:} The methods that maximize the margin in Euclidean or angular spaces mentioned above have the shortcomings in the ways that the objective loss functions include many terms that need to be weighted, the class acceptance regions are not compact, or they need additional hard-mining algorithms. In this study, we propose a simple yet effective method that does not have these limitations. Our proposed method maximizes the margin in both the Euclidean and angular spaces. To the best of our knowledge, our proposed method is the first method that maximizes the margin in both spaces. To accomplish this goal, we train a deep neural network that enforces the samples to gather in the vicinity of the class-specific centers that lie on the boundary of a hypersphere. Each class is represented with a single center and the distances between the class centers are equivalent. This corresponds to selection of class centers from the vertices of a regular simplex inscribed in a hypersphere. Both the Euclidean distances and angular distances between class centers are equivalent to each other.

Our proposed method has many advantages over other margin maximizing deep neural network classifiers. These advantages can be summarized as follows:
\begin{itemize}
    \item The proposed loss function does not have any hyperparameter that must be fixed for classical classification problems, therefore it is extremely easy for the users. For open set recognition, the user has to set two parameters if the background class samples are used for learning.
    \item The proposed method returns compact and interpretable acceptance regions for each class, thus it is very suitable for open set recognition problems.  
    \item The distances between the samples and their corresponding centers are minimized independently of each other, thus the proposed method also works well for imbalanced datasets.
\end{itemize}
In contrast, there is only one limitation of the proposed method: The dimension of the CNN features must be larger than or equal to the total number of classes minus 1. To overcome this limitation, we introduced Dimension Augmentation Module (DAM) as explained below.

\begin{figure*}[t]
		\begin{center}
				\includegraphics[width=\textwidth]{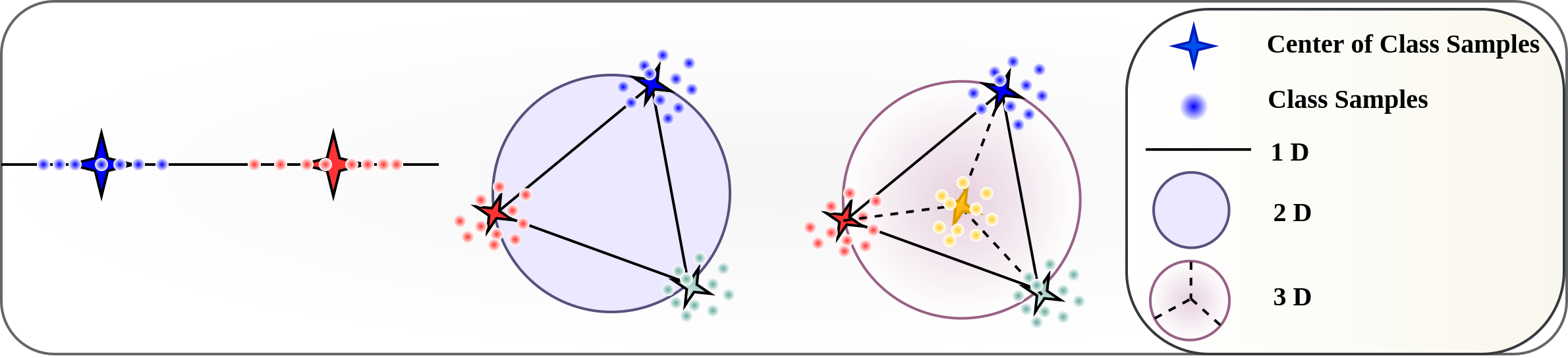}
				\caption{In the proposed method, class samples are enforced to lie closer to the class-specific centers representing them, and the class centers are located on the boundary of a hypersphere. All the distances between the class centers are equivalent, thus there is no need to tune any margin term. The class centers form the vertices of a regular simplex inscribed in a hypersphere. Therefore, to separate $C$ different classes, the dimensionality of the feature space must be at least $C-1$. The figure on the left shows separation of 2 classes in 1-D space, the middle figure depicts the separation of 3 classes in 2-D space, and the figure on the right illustrates the separation of 4 classes in 3-D space. For all cases, the centers are chosen from a regular $C-$simplex.}
					\label{fig:1}
		\end{center}		
\end{figure*}

\section{Method}
\subsection{Motivation}
In this study, we propose a simple yet effective deep neural network classifier that maximizes the margin in both Euclidean and angular spaces. To this end, we introduce a novel classification loss function that enforces the samples to compactly cluster around the class-specific centers that are selected from the outer boundaries of a hypersphere. The Euclidean distances and angles between the centers are equivalent. This is illustrated in Fig. \ref{fig:1}. In this figure, the centers representing the classes are denoted by the star symbols whereas the class samples are represented with circles having different colors based on the class memberships. As seen in the figure, all pair-wise distances between the class centers are equivalent, and class centers are located on the boundary of a hypersphere. Moreover, if the hypersphere center is set to the origin, then the angles between the class centers are also same, and the lengths of the centers are equivalent, i.e, $\left\|\ssc_i\right\|=u$, ($u$ is the length of the center vectors). After learning stage, if the class samples are compactly clustered around the centers representing them, we can classify the data samples based on the Euclidean or angular distances from the class centers. Both distances yield the same results if the hypersphere center is set to the origin.

At this point, the question of whether enforcing data samples to lie around the simplex vertices is appropriate or not comes to mind. In fact, high-dimensional spaces are quite different than the low dimensional spaces, and there are many studies showing that the data samples lie on the boundary of a hypersphere when the feature dimensionality, $d$, is high and the number of samples, $n$, is small. For example, \cite{R51} theoretically show that the high-dimensional spaces are mostly empty and data concentrate on the outside of a shell (on the outer boundary of a hypersphere). They also show that as the number of dimensions increases, the shell increases its distance from the origin. More precisely, the data samples lie near the outer surface of a growing hypersphere in high-dimensional spaces (therefore setting the hypersphere radius to 1 as in \cite{RNN4} is not suitable for high-dimensional spaces). In a more recent study, \cite{R52} explicitly show that the data samples lie at the vertices of a regular simplex in high-dimensional spaces. These two studies are not contradictory and they support each other since we can always inscribe a regular simplex in a hypersphere as seen in Fig. \ref{fig:1}. In addition to these studies, \cite{R53,R54} show that the eigenvectors of the Laplacian matrices (the matrices computed by operating on similarity matrices in spectral clustering analysis) form a simplex structure, and they use the vertices of resulting simplex for clustering of data samples. In other words, they prove that when the data samples are mapped to Laplacian eigenspace, they concentrate on the vertices of a simplex structure. These studies are also complementary to the studies showing that the high-dimensional data samples lie on the boundary of a growing hypersphere. It is because, as proved in \cite{R55}, NCuts (Normalized Cuts) \cite{R56} clustering algorithm, which is presented as a spectral relaxation of a graph cut problem, maps the data samples onto an infinite-dimensional feature space. Therefore, these data samples naturally concentrate on the vertices of a regular simplex due to the high-dimensionality of the feature space. 
A recent study \cite{RNN5} also reveals that the lengths of the vectors of the class means (after centering by their global mean) converge to the same length and the angles between pair-wise center vectors become equal during the last training stages (it is called terminal phase of the training in the study) of the deep neural network networks using linear classifiers at the end. They also demonstrate that the within-class scatter also converges to zero indicating that the class-specific samples gather around their corresponding class center. More precisely, they show that the samples of different classes cluster around the class centers forming the vertices for a regular simplex (as we proposed in this study) at the last stages of the learning process. Furthermore, they also demonstrate that the deep neural network classifier eventually converges to choosing whichever class has the nearest mean. However, the study is not complete in the sense that  they do not consider the cases when the dimension is smaller than the number of classes so that it is impossible to fit the class centers to the vertices of a regular simplex.

\subsection{Maximizing Margin in Euclidean and Angular Spaces}
In the proposed method, we map the class samples to compactly cluster around the class centers chosen from the vertices of a regular simplex. All the pair-wise distances between the selected class centers are equivalent. Assume that there are $C$ classes in our data set. In this case, we first need to create a $C$-simplex (some researchers call it $C-1$ simplex considering the feature dimension, but we will prefer $C$-simplex definition). The vertices of a regular simplex inscribed in a hypersphere with radius 1 can be defined as follows:

\begin{equation}
 \vv_j=\left\{\begin{array}{ll}
	(C-1)^{-1/2}\OO, &  j=1, \\
	\kappa \OO+\eta\ee_{j-1}, & 2\leq j \leq C,
\end{array} \right.
\end{equation}
where,
\begin{equation}
\kappa=-\frac{1+\sqrt{C}}{(C-1)^{3/2}},  \eta=\sqrt{\frac{C}{C-1}}.
\end{equation}
Here, $\OO$ is an appropriate sized vector whose elements are all 1, $\ee_j$ is the natural basis vector in which the $j-$th entry is 1 and all other entries are 0. Such a $C-$simplex is in fact a $C-$dimensional polyhedron where the distances between the vertices are equivalent. It must be noted that the distances between the vertices do not change even if the simplex is rotated or translated. But, the dimension of the feature space must be at least $C-1$ in order to define such a regular $C-$simplex. Next, we must define the radius, $u$, of the hypersphere. This term is similar to the scaling parameter used in methods such as ArcFace \cite{R18}, CosFace \cite{R16}, etc. that maximize the margin in angular spaces. As the dimension increases, it must be also increased since the studies \cite{R51} show that the hypersphere whose outer shells include the data also grows as the dimension is increased. We set $u=64$ as in ArcFace method. Then, we set the class centers that will represent the classes as,
\begin{equation}
\ssc_j=u\vv_j, \:\:\:  j=1,...,C.
\end{equation}
The order of selection of centers does not matter since the distances among all centers are equivalent. 
Now, let us consider that the deep neural network features of training samples are given in the form $(\ff_i,y_i)$, $i=1,\ldots,n$, $\ff_i \in \R^d$, $y_{i} \in \left\{j\right\}$ where ${j=1,...,C}$. Here, $C$ is the total number of known classes, and we assume that the feature dimension $d$ is larger than or equal to $C-1$, i.e., $d\geq C-1$.
In this case, the loss function of the proposed method can be written as,
\begin{equation}
\mathcal{L} = \frac{1}{n}\sum_{i=1}^{n}\left\|\ff_i-\ssc_{y_i}\right\|^2.
\label{eq:lfunc}
\end{equation}
The loss function includes a single term that aims to minimize the within-class variations by minimizing the distances between the samples and their corresponding class centers which are set to the vertices of a regular simplex. There is no need another loss term for the between-class separation since the selected centers have the maximum possible Euclidean and angular distances among them. As a result, there is no hyperparameter that must be fixed, and the proposed method is extremely easy for the users. Moreover, the data samples compactly cluster around their class centers, therefore the proposed method returns compact acceptance regions for classes, which is crucial for the success of the open set recognition. We call the resulting methods as \textit{Deep Simplex Classifier (DSC)}.

\subsection{Including Background Class for Open Set Recognition}
In open set recognition problems, novel classes (ones not seen during training) may occur at test time, and the goal is to classify the known class samples correctly while rejecting the unknown class samples \cite{R26}. Earlier open set recognition methods only used the known class samples during training. However, more recent studies \cite{RR33,RR37,R57} revealed that using the background dataset that includes the samples that come from the classes that are different from the known classes greatly improves the accuracies. Let us represent the deep neural network features of the background samples by $\ff_k \in \R^d$, $k=1,...,K$.
In order to incorporate the background samples, we add an additional loss term that pushes the background samples away from the known class centers as follows:
\begin{equation}
\Scale[0.70]{\mathcal{L} = \frac{1}{n}\sum_{i=1}^{n}\left\|\ff_i-\ssc_{y_i}\right\|^2+
\lambda \sum_{i=1}^{n} \sum_{k=1}^{K} \argmax\left(0,m+\left\|\ff_i-\ssc_{y_i}\right\|^2-\left\|\ff_k-\ssc_{y_i}\right\|^2 \right)},
\end{equation}
where $m$ is the selected threshold, and $\lambda$ is the weighting term. The second loss term enforces the distances between the known class samples and their corresponding class centers to be smaller than the distances between the background class samples and the known class centers by at least a selected margin, $m$. In contrast to our first proposed loss function, this loss function includes two terms that must be set by the users. But, this is necessary only if we use the background class samples. 

\begin{figure}[tbh]
		\begin{center}
				\includegraphics[width=\columnwidth]{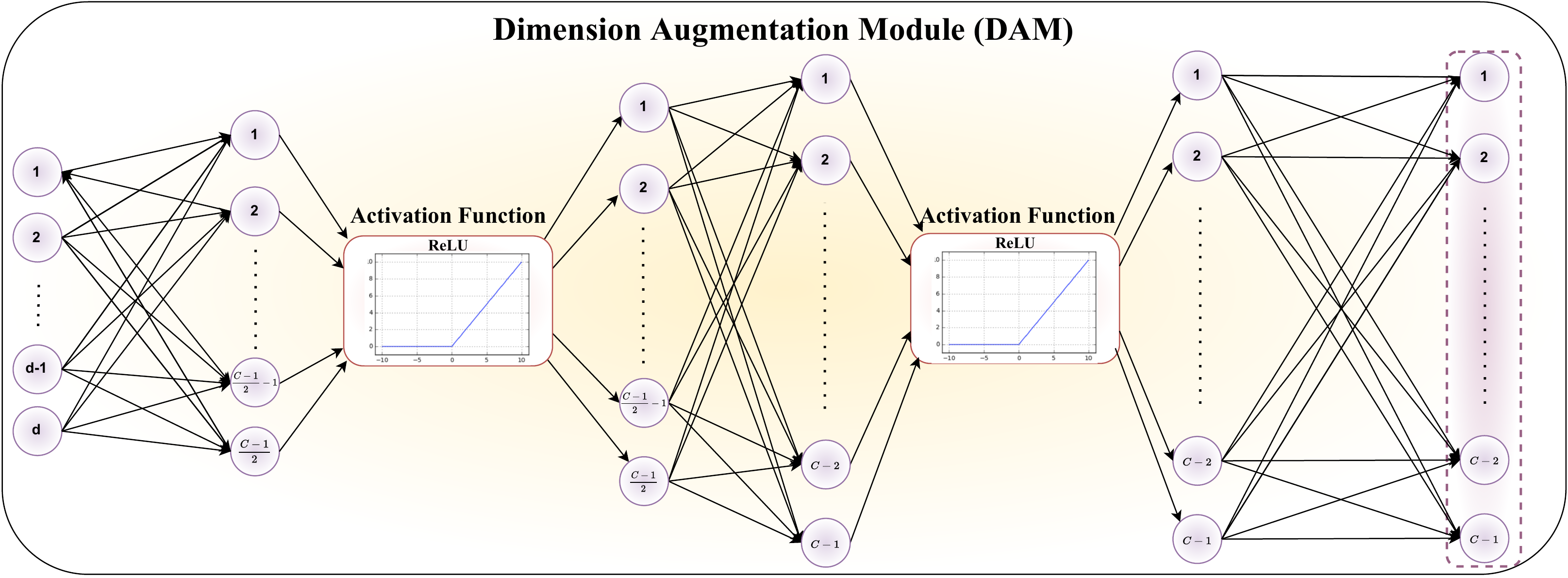}
				\caption{The plug and play module that will be used for increasing feature dimension. It maps $d-$dimensional feature vectors onto a much higher $(C-1)-$ dimensional space.}
					\label{fig:2}
		\end{center}		
\end{figure}

\subsection{Dimension Augmentation Module (DAM)}
The major limitation of the proposed method is the restriction that the dimension of the feature space must be larger than or equal to $C-1$, i.e., $d\geq C-1$. A similar restriction exists in \cite{RNN4}, and their proposed method requires $d\geq C$ since they choose the class centers as the standard basis vectors of $C$-dimensional space as opposed to our proposed method that selects the centers from the vertices of a regular simplex. The typical feature dimension size returned by the classical deep neural network classifiers is 2048 or 4096. In this case, the number of classes in our training set cannot exceed 2049 or 4097. However, the number of classes can be larger than these values for some classification tasks, and we cannot use the proposed method in such cases. There are basically two procedures to solve this problem: As a first solution, we can use a method similar to \cite{R58} that returns more centers where the distances between centers are approximately equivalent. In this case, the number of centers is increased to $2d+4$ for $d-$dimensional feature spaces. As a second and a more complete solution, we introduce a plug and play module called Dimension Augmentation Module (DAM) that increases the feature dimension size to any desired value. The module is visualized in Fig. \ref{fig:2}, and it includes two fully connected layers supported with activation functions. The first fully connected layer maps the $d-$dimensional feature space onto a higher $C-1$ dimensional space. Then, we apply ReLU (Rectified Linear Unit) activation functions followed by the second fully connected layer. This is similar to kernel mapping idea used in kernel methods \cite{R59,R60} in the spirit with the exception that we explicitly map the data to higher dimensional feature space as in \cite{RR38,RR45}. 
It should be noted that \cite{RNN4} proposed to use a fully connected layer alone for increasing the dimensionality of the feature space. However, this is a mistake since a fully connected layer just uses the linear combination of existing features and the resulting space has the same dimensionality as in the original feature space. They have to use activation functions to introduce non-linearity and increase the dimension. 

\section{Experiments}
\subsection{Illustrations and Ablation Studies}
Here, we first conducted some experiments to visualize the embedding spaces returned by the various loss functions using the vertices of the regular simplex. For this illustration experiment, we designed a deep neural network where the output of the last hidden layer is set to 2 for visualizing the learned features. %This allows to plot the learned feature vectors in 2D space. 
As training data, we selected 3 classes from the Cifar-10 dataset.
We would like to point out that we can use different loss functions in addition to our default loss function given in (\ref{eq:lfunc}) once we determine the vertices of the simplex that will represent the classes. To this end, we used two other loss functions: The first one is the hinge loss that minimizes the distances between the samples and their corresponding class center if the distance is larger than a selected threshold,
\begin{equation}
\mathcal{L}_{hinge} = \frac{1}{n}\sum_{i=1}^{n} \argmax\left(0, \left\| \ff_i -\ssc_{y_i}\right\|^2-m\right).
\end{equation}
This loss function does not minimize the distances between the samples and their corresponding centers if the distances are already smaller than the selected threshold, $m$. This way class-specific samples are collected in a hypersphere with radius, $m$. For the second loss function, we used the variant of the softmax loss function where the weights are fixed to the simplex vertices as in,
\begin{equation}
\mathcal{L}_{softmax} = -\frac{1}{n}\sum_{i=1}^{n} \log \frac{e^{\ssc_{y_i}^\top \ff_i+b_{y_i}}}{\sum_{j=1}^C e^{\ssc_{j}^\top \ff_i+b_{j}}}
\end{equation}
For the softmax loss, we fix the classifier weights to the pre-defined class centers and we only update features of the samples by using back-propagation. We set the hypersphere radius to, $u=5$, since this is a simple dataset.

\begin{figure}[t]
\centering
\begin{tabular}[h]{@{}ccc@{}}
\includegraphics[width=0.3\columnwidth]{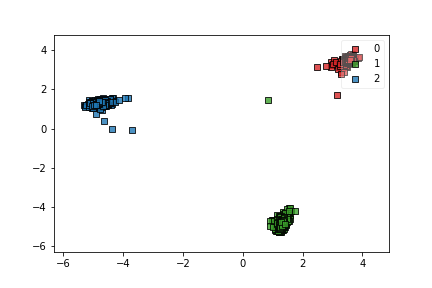} &
\includegraphics[width=0.3\columnwidth]{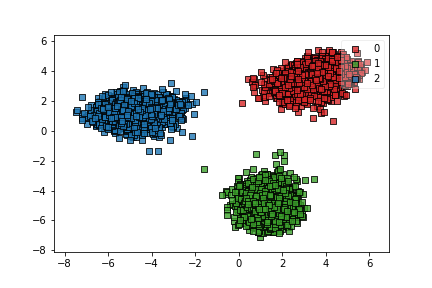}  &
\includegraphics[width=0.3\columnwidth]{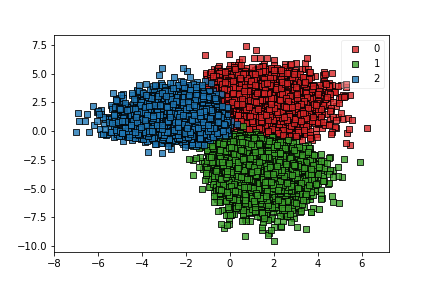}  \\
(a) & (b) & (c) \\
\end{tabular}
\caption{Learned feature representations of image samples: (a) the embeddings returned by the proposed method trained with the default loss function given in (\ref{eq:lfunc}), (b) the embeddings returned by the proposed method trained with the hinge loss, (c) the embeddings returned by the proposed method trained with the softmax loss function. }
\label{fig:3}
\end{figure}

The embeddings returned by the deep neural networks using different loss functions are plotted in Fig. \ref{fig:3}. The first figure on the left is obtained by our default loss function that does not need any parameter selection. All data samples are compactly clustered around their class means as expected. The second loss function using the hinge loss returns spherical distributions based on the selected margin, $m$, and the classes are still separable by a margin. In contrast, when the softmax is used with the simplex vertices, the data samples are very close and they overlap since there is no margin among the classes. Therefore, our default loss function seems to be the best choice among all tested variants since it does not need fixing any parameter and returns compact class regions.

\subsection{Open Set Recognition Experiments}
For open set recognition, we need to split the datasets into \textit{known} and \textit{unknown} classes. To this end, we used the common standard settings that  are also applied for testing other recent open set recognition methods. The details of each dataset and its open set recognition setting are given below. By following the standard protocol, random splitting of each dataset into known and unknown classes is repeated 5 times, and the final accuracies are averages of the results obtained in each trial. 

We compared our proposed method, Deep Simplex Classifier (DSC), to other state-of-the-art open set recognition methods including Softmax, OpenMax \cite{R26}, C2AE \cite{R39}, CAC \cite{RR37}, CPN \cite{R44}, OSRCI \cite{R38}, CROSR \cite{RR39}, RPL \cite{RN39}, %ARPL \cite{RR40}, 
Objecttosphere \cite{R33}, and Generative-Discriminative Feature Representations (GDFRs) \cite{R47} methods. 
 We used the same network architecture used in \cite{R38} as our backbone network for all datasets with the exception of TinyImageNet dataset, where we preferred a deeper Resnet-50 architecture for this dataset. We started the training from completely random weights (without any fine-tuning). Therefore, our proposed method is directly comparable to the published results in \cite{R38} for majority of the tested datasets.

\begin{table*}[t]
\begin{center}
\caption{AUC Scores (\%) of open set recognition methods on tested datasets ($n.r.$ stands for not reported).}
\label{table:opensetres}
\vspace{-0.05cm}
\resizebox{\textwidth}{!}
{
  \begin{tabular}[t] {|l|c|c|c|c|c|c|c|} \hline
    \textbf{\small {Methods}}  & \textbf{Mnist} & \textbf{Cifar10} & \textbf{SVHN}  & \textbf{Cifar+10} & \textbf{Cifar+50} & \textbf{TinyImageNet}\\ \hline\hline
    DSC (Ours) & $\mathbf{99.6}\pm 0.1$ & $93.8\pm 0.3$ & $95.3\pm 0.8$ & $\mathbf{99.1}\pm 0.2$ & $\mathbf{98.4}\pm 0.3$ & $\mathbf{82.5}\pm 1.8$  \\ \hline
		Softmax & $97.8\pm 0.2$ & $67.7\pm 3.2$ & $88.6\pm 0.6$ & $81.6\pm n.r.$ & $80.5\pm \pm n.r.$ & $57.7\pm n.r.$   \\ \hline
		OpenMax & $98.1\pm 0.2$ & $69.5\pm 3.2$ & $89.4\pm 0.8$ & $81.7\pm n.r.$ & $79.6\pm n.r.$ & $57.6\pm n.r.$   \\ \hline
		G-OpenMax & $98.4\pm 0.1$ & $67.5\pm 3.5$ & $89.6\pm 0.6$ & $82.7\pm n.r.$ & $81.9\pm n.r.$ & $58.0\pm n.r.$   \\ \hline
		C2AE & $98.9\pm 0.2$ & $89.5\pm 0.9$ & $92.2\pm 0.9$ & $95.5\pm 0.6$ & $93.7\pm 0.4$ & $74.8\pm 0.5$  \\ \hline
		CAC & $99.1\pm 0.5$ & $80.1\pm 3.0$ & $94.1\pm 0.7$ & $87.7\pm1.2$ & $87.0\pm 0.0$ & $76.0\pm 1.5$  \\ \hline
		CPN & $99.0\pm 0.2$ & $82.8\pm 2.1$ & $92.6\pm 0.6$ & $88.1\pm n.r.$ & $87.9\pm n.r.$ & $63.9\pm n.r.$  \\ \hline
		OSRCI & $98.8\pm 0.1$ & $69.9\pm 2.9$ & $91.0\pm 0.6$ & $83.8\pm n.r.$ & $82.7\pm -$ & $58.6\pm n.r.$   \\ \hline
		CROSR & $99.1\pm n.r.$ & $88.3\pm n.r.$ & $89.9\pm n.r.$ & $91.2\pm n.r.$ & $90.5\pm n.r.$ & $58.9\pm n.r.$   \\ \hline
		RPL & $98.9\pm 0.1$ & $82.7\pm 1.4$ & $93.4\pm 0.5$ & $84.2\pm 1.0$ & $83.2\pm 0.7$ & $68.8\pm 1.4$   \\ \hline
		GDFRs & $n.r.$ & $83.1\pm 3.9$ & $\mathbf{95.5}\pm 1.8$ & $92.8\pm 0.2$ & $92.6\pm 0.0$ & $64.7\pm 1.2$   \\ \hline
		Objecttosphere & $n.r.$ & $\mathbf{94.2}\pm n.r.$ & $91.4\pm n.r.$ & $94.5\pm n.r.$ & $94.4\pm n.r.$ & $75.5\pm n.r.$   \\ \hline
		%Energy & $--$ & $93.3$ & $89.4$ & $93.6$ & $93.7$ & $72.7$   \\ \hline
		\end{tabular}}
\end{center}
\vspace{-0.4cm}
\end{table*}

\subsubsection{Datasets}
\noindent{\bf Mnist, Cifar10, SVHN:} By using the standard setting, Mnist, Cifar10, and SVHN datasets are split randomly into 6 known and 4 unknown classes. We used 80 Million Tiny Images dataset \cite{R61} as the background class. \\
\noindent{\bf Cifar+10, Cifar+50:} For Cifar+$N$ experiments, we use 4 randomly selected classes from Cifar10 dataset for training, and $N$ non-overlapping classes chosen from Cifar100 dataset are used as unknown classes as in \cite{R44,RR37,RR39,RN39}. We used 80 Million Tiny Images dataset \cite{R61} as the background class. \\
\noindent{\bf TinyImageNet:} For TinyImageNet \cite{R62} experiments, we randomly selected 20 classes as known classes and 180 classes as unknown classes by following the standard setting. We used 80 Million Tiny Images dataset \cite{R61} as the background class.

\subsubsection{Results}
For open set recognition, Area Under the ROC curve (AUC) scores are used for measuring the detection of performance of the unknown samples. In addition, we also report the closed-set accuracy for measuring the classification performance on known data by ignoring the unknown samples as in \cite{R44,R38} (these results are given in Appendix). AUC scores are given in Table \ref{table:opensetres}. As seen in the table, our proposed method achieves the best accuracies on all datasets with the exception of Cifar 10 and SVHN datasets. The performance difference is very significant especially on Cifar+10, Cifar+50 and TinyImageNet datasets.

\subsection{Closed Set Recognition Experiments}
\subsubsection{Experiments on Moderate Sized Datasets}
Here, we conducted closed set recognition experiments on moderate sized datasets. Our proposed method did not need DAM since the feature dimension is much larger than the number of classes in the training set for these experiments. We compared our results to the methods that maximize the margin in Euclidean or angular spaces. We implemented the compared methods by using provided source codes by their authors, and we used the ResNet-18 architecture \cite{R22} as backbone for all tested methods. Therefore, our results are directly comparable. 

\begin{table}[tbh]
\begin{center}
\caption{Classification accuracies (\%) on moderate sized datasets.}
\label{moderateres}
%\small
  \begin{tabular}[tbh] {|l|c|c|c|} \hline
    \textbf{Methods}  & \textbf{Mnist} & \textbf{Cifar-10} & \textbf{Cifar-100} \\ \hline\hline
		 DSC (Ours) & $\mathbf{99.7}$ &  $\mathbf{95.9}$ & $\mathbf{79.5}$  \\ \hline
		Softmax & $99.4$ &  $94.4$ & $75.3$  \\ \hline
		Center Loss & $\mathbf{99.7}$ &  $94.2$ & $76.1$  \\ \hline
		ArcFace & $\mathbf{99.7}$ &  $94.8$ & $75.7$  \\ \hline
		CosFace & $\mathbf{99.7}$ &  $95.0$ & $75.8$  \\ \hline
		SphereFace & $\mathbf{99.7}$ &  $94.7$ & $75.1$  \\ \hline
			\end{tabular}
   \end{center}
\vspace{-0.2cm}
\end{table}

Classification accuracies are given in Table \ref{moderateres}. For Mnist datasets, majority of the tested methods yield the same accuracy, but our proposed DSC method outperforms all tested methods on the Cifar-10 and Cifar-100 datasets. The performance difference is significant especially on the Cifar-100 dataset. These results verify the superiority of the margin maximization in both Euclidean and angular spaces. Achieving the best accuracies is encouraging, because our proposed method is very simple and does not need any parameter tuning, yet it outperforms more complex methods.  

\subsubsection{Experiments on Large-Scale Datasets}
For all face verification tests, we used the same network trained on large-scale face dataset by following the standard setting. To this end, we trained the proposed classifier on MS1MV2 dataset \cite{R18}, which is a cleaned version of MS-Celeb-1M dataset \cite{R32}. This dataset includes approximately 85.7K individuals. We removed the classes including less than 100 samples, which left us approximately 18.6K individuals for training. The number of classes is much larger than the feature dimension, $d=2048$, thus we used DAM to increase the CNN feature dimension. The ResNet-101 architecture is used as backbone. Once the network is trained, we used the resulting architecture to extract deep CNN features of the face images coming from the test datasets. 

As test datasets, we used Labeled Faces in the Wild (LFW) \cite{RHN2}, Cross-Age LFW (CALFW) \cite{RHN3}, Cross-Pose LFW (CPLFW) \cite{RHN4}, Celebrities in Frontal-Profile data set (CFP-FP) \cite{RHN6}  and AgeDB \cite{RHN6}. %LFW includes 13,233 color face images of 5,749 subjects collected from the web. CALFW is constructed by adding 3,000 positive face pairs with age gaps to LFW dataset. CPLFW dataset is collected to test the verification performance when cross-pose face images of same individuals are used for comparison. CFP-FP dataset contains 10 frontal and 4 profile images of 500 individuals. Similar to LFW, 10 splits are created and each split contains 350 same and 350 not-same pairs. AgeDB dataset is collected to compare methods on age-invariant face verification, age estimation and face age progression “in-the-wild”. It contains 16,488 images of various famous people. Every image is annotated with respect to the identity, age and gender attribute. There exists a total of 568 distinct subjects. The average number of images per subject is 29.
We evaluated the proposed methods by following the standard protocol of unrestricted with labeled outside data \cite{RHN2}, and report the results by using 6,000 pair testing images on LFW, CALFW, CPLFW, and AgeDB. However, 7,000 pairs of testing images are used for CFP-FP by following the standard setting. The results are given in Table \ref{verification_res}. As seen in the results, the proposed method using DAM outperforms the classifiers using softmax and Center loss, but accuracies are lower than the recent state-of-the-art methods. These results indicate that the DAM solves the dimension problem partially, but it must be revised for obtaining better accuracies.

\begin{table}[tbh]
 	\begin{center}
 		\renewcommand{\arraystretch}{1.2}
 		\caption{Verification rates (\%) on different datasets.}
 		%\small
 		\label{verification_res}
 		\def\B#1{\textbf{#1}}
 		\tabcolsep 2.5pt 
 		\begin{tabular}[tb] {|l|c|c|c|c|c|} \hline
		 {\textbf{Method}}  &  {\textbf{LFW}}  & {\textbf{CALFW}} & {\textbf{CPLFW}} & {\textbf{CFP}}  & {\textbf{AgeDB}}\\  \hline	
			DSC & $99.6$  & $91.3$ & $90.3$ & $94.3$ & $96.0$   \\ \hline
			VGGFace2  & $99.4$ & $90.6$ & $84.0$ & $--$ & $--$\\ \hline
			Center Loss  & $99.3$ & $85.5$ & $77.5$ & $--$ & $--$   \\ \hline
			ArcFace (ResNet-101) & $\mathbf{99.8}$  & $\mathbf{95.5}$  & $\mathbf{92.1}$ & $95.6$ & $--$ \\ \hline 
			%GroupFace (ResNet-100) \cite{RR23}  & $99.8$ & $98.6$ & $96.2$ & $93.2$ & $98.3$\\ \hline
			CosFace & $99.7$ & $93.3$ & $\mathbf{92.1}$ & $--$ & $\mathbf{97.7}$ \\ \hline
			SphereFace  & $99.4$ &  $93.3$ & $\mathbf{92.1}$ & $94.4$ &  $\mathbf{97.7}$ \\ \hline
			\end{tabular}
 	\end{center}
  \end{table}

\section{Summary and Conclusion}
In this paper, we proposed a simple and effective deep neural network classifier that maximizes the margin in both the Euclidean and angular spaces. The proposed method returns embeddings where the class-specific samples lie in the vicinity of the class centers chosen from the vertices of a regular simplex. The proposed method is very simple in the sense that there is no parameter that must be fixed for classical closed set recognition settings. Despite its simplicity, the proposed method achieves the state-of-the-art accuracies on open set recognition problems since the samples of unknown classes are easily rejected by using the distances from the class-specific centers. Moreover, our proposed method also outperformed other state-of-the-art classification methods on closed set recognition setting when moderate sized datasets are used. The proposed method has a limitation regarding learning in large-scale datasets. We introduced DAM in order to solve this problem. Although DAM partially solved the existing problem, we could not get state-of-the-art accuracies on large-scale face recognition problems. As a future work, we are planning to improve DAM by changing its architecture and activation functions.

\smallskip\noindent{\bf Acknowledgments:} This work was
funded in part by the Scientific and Technological Research Council of
Turkey (TUB\.{I}TAK) under Grant number EEEAG-121E390. % \balance

%\section*{References}
\bibliography{myreferences}{}
\bibliographystyle{unsrt}

\section*{Appendix}
Here, we first explain the implementation details of the proposed deep neural network classifier, and give the parameters used for the utilized deep neural network classifier architecture. This is followed by a discussion for other alternatives to circumvent the dimension restriction. Then, we reported the closed-set accuracies of tested methods on open set recognition datasets. Finally, we conduct experiments verifying that our proposed method returns semantically related feature representations and it is more robust against imbalanced datasets.

\subsection{Implementation Details}
For open set recognition, we used the same network architecture used in \cite{R38} as our backbone network for all datasets with the exception of TinyImageNet dataset, where we preferred a deeper Resnet-50 architecture for this dataset. The learning rate is set to $0.1$. For open set recognition experiments, we set $\lambda=\frac{1}{2\times batch\_size^2}$, and $m=u/2$, where $u$ is the hypersphere radius. 

We do not need $\lambda$ and $m$ parameters for closed set recognition. For closed-set recognition experiments, we used the ResNet-18 architecture as backbone for moderate sized datasets, and the ResNet-101 architecture is used for large-scale face recognition dataset. For updating network weights, we used Adam optimization strategy for large-scale face recognition whereas SGD (stochastic gradient descent) is used for moderate size datasets. The learning rate is set to $10^{-3}$ for face recognition and to $0.5$ for moderate sized datasets.

Regarding the scale parameter $u$, we have conducted experiments by selecting different values. Experiments verify that the selection of $u$ is not very important as long as it is not fixed to small values such as 1. Theoretically, the data samples lie on the surface of a growing hypersphere as the dimension increases. For smaller dimensions, we can choose smaller values of $u$ as we did for illustrations experiments (we fixed $u$ to 5 for 2-dimensional inputs). But, for larger dimensions we need higher values. Also, after some value, increasing $u$ value does not change the results much. The accuracies that are obtained for Cifar-100 dataset for various $u$ values are given in Table \ref{uvalue}.

\begin{table}[tbh]
\begin{center}
\caption{Classification accuracies (\%) for different $u$ values on Cifar-100 dataset.}
\label{uvalue}
%\small
  \begin{tabular}[tbh] {|l|c|} \hline
    \textbf{$u$ values}  & \textbf{Accuracies (\%)} \\ \hline\hline
		 $u=32$ & $76.2$   \\ \hline
		$u=64$ & $79.5$   \\ \hline
		$u=100$ & $79.4$   \\ \hline
		$u=150$ & $79.9$   \\ \hline
		$u=200$ & $79.0$   \\ \hline
			\end{tabular}
   \end{center}
\vspace{-0.2cm}
\end{table}

The threshold, $m$, parameter is only used for open set recognition problems. Moreover, we did not have any trouble for fixing it since our centers are fixed to certain positions. We already know the distances between the class centers chosen as simplex vertices. All distances are equal, we simply checked the largest intra-class distances within classes and determined a margin based on this. Setting margin term to half of the radius worked well for all cases. For all experiments, we did not fine-tune our classification network from a pre-trained network and started the network weights from scratch by initializing with random weights which is the common practice used for initializing network weights.

\subsection{Discussion on the Feature Dimension Restriction}
In order to circumvent the feature dimension restriction, we used DAM in the proposed method. The main idea was to design a plug and play module that can be used with any desired deep CNN architecture without any changes. DAM module partially solved our problem and yielded satisfactory accuracies closer to the state-of-the-art. We can provide a better solution by designing our own architecture instead of using our proposed plug and play DAM module. To this end, we can avoid the fully connected layers that are used for dimension reduction in the last layers of deep CNNs. For example, in the ResNet architectures, the dimension of the feature space is 25088 just before fully connected layers, and it is reduced to 512 after fully connected layers. We can avoid the last fully connected layers and use high-dimensional outputs of these earlier layers. For example, using 25088 dimensional feature space is enough for training the large-scale MS1MV2 dataset we used in our tests without any need for dimension increase. 

\subsection{Closed-Set Accuracies on Open Set Recognition Datasets}
Closed-set accuracies of the open-set recognition methods are given in Table \ref{closed_res}. Our proposed method also obtains the best closed-set accuracies among the tested methods with the exception of SVHN dataset. This clearly shows that the proposed method is very successful both at the rejection of the unknown samples and classification of the known samples correctly. 

\begin{table}[tbh]
\begin{center}
\caption{Closed-Set accuracies (\%) of open set recognition methods on tested datasets.}
\label{closed_res}
%\small
\resizebox{\columnwidth}{!}
{
  \begin{tabular}[tbh] {|l|c|c|c|c|c|c|c|} \hline
    \textbf{\small {Methods}}  & \textbf{Mnist} & \textbf{Cifar10} & \textbf{SVHN}  & \textbf{Cifar+10} & \textbf{Cifar+50} & \textbf{TinyImageNet}\\ \hline\hline
    DSC (Ours) & $\mathbf{99.8}\pm 0.1$ & $\mathbf{96.1}\pm 1.4$ & $96.5\pm 0.3$ & $\mathbf{97.6}\pm 0.5$ & $\mathbf{97.9}\pm 0.5$   & $\mathbf{83.3}\pm 2.2$    \\ \hline
		Softmax & $99.5\pm 0.2$ & $80.1\pm 3.2$ & $94.7\pm 0.6$ & $n.r.$ & $n.r.$ & $n.r.$   \\ \hline
		OpenMax & $99.5\pm 0.2$ & $80.1\pm 3.2$ & $94.7\pm 0.6$ & $n.r.$ & $n.r.$ & $n.r.$   \\ \hline
		G-OpenMax & $99.6\pm 0.1$ & $81.6\pm 3.5$ & $94.8 \pm 0.8$ & $n.r.$ & $n.r.$ & $n.r.$ \\ \hline
		CPN & $99.7\pm 0.1$ & $92.9\pm 1.2$ & $\mathbf{96.7} \pm 0.4$ & $n.r.$ & $n.r.$ & $n.r.$ \\ \hline
		OSRCI & $99.6\pm 0.1$ & $82.1\pm 2.9$ & $95.1 \pm 0.6$ & $n.r.$ & $n.r.$ & $n.r.$ \\ \hline
		CROSR & $99.2\pm 0.1$ & $93.0\pm 2.5$ & $94.5 \pm 0.5$ & $n.r.$ & $n.r.$ & $n.r.$ \\ \hline
			\end{tabular}}
\end{center}
\vspace{-0.3cm}
\end{table}

\subsection{Semantically Related Feature Embeddings}

\begin{figure}[tbh]
		\begin{center}
				\includegraphics[width=0.8\columnwidth]{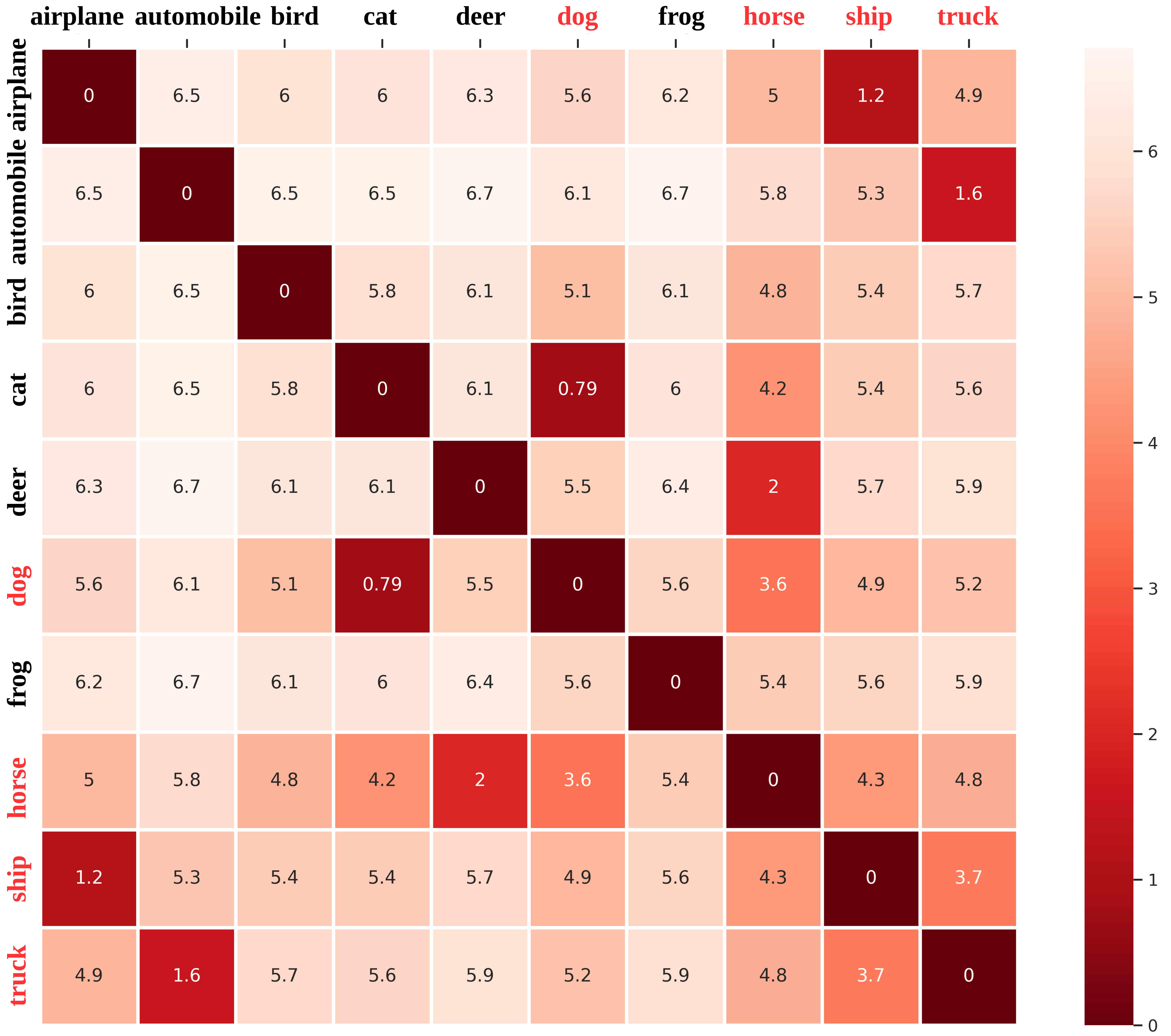}
				\caption{The distance matrix computed by using the centers of the testing classes. The four classes that are not used in training are closer to their semantically related classes in the learned embedding space.}
					\label{fig:4}
		\end{center}		
\end{figure}

We also conducted experiments to see if the proposed method returns meaningful feature embeddings where the semantically and visually similar classes lie close to each other in open set recognition settings. It should be noted that the semantic relationships are not preserved for the training classes since the Euclidean and angular distances between the class centers are equivalent. However, if the proposed method returns good CNN features, we expect the samples belonging to classes not used in training to lie closer to their semantically related training classes. To verify this, we trained our proposed method by using 6 classes from the Cifar-10 dataset: airplane, automobile, bird, cat, deer, and frog. Then, we extracted the CNN features of all testing data coming from 10 classes by using the trained network. Then, we computed the average CNN feature vector of each class, and computed the distances between them. Fig. \ref{fig:4} illustrates the computed distances between the centers. The distances between the classes used for training are similar and they change between 5.8 and 6.7. The four classes, the dog, horse, ship, and truck classes, that are not used for training are represented with red color in the figure. As seen in the figure, the dog class is closest to its semantically similar cat class, the truck class is closer to its semantically similar automobile class, the horse class is closest to the deer class, and the ship class is closer to the visually similar airplane class (since the backgrounds - blue sky and sea - are mostly similar for these two classes). This clearly shows that the proposed method returns semantically meaningful embeddings.   

\subsection{Experiments on Imbalanced Datasets}
\begin{figure}[tbh]
\centering
\begin{tabular}[h]{@{}cc@{}}
\includegraphics[width=0.5\columnwidth]{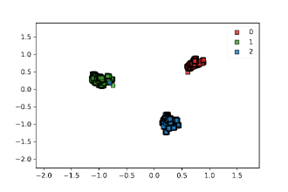} &
\includegraphics[width=0.5\columnwidth]{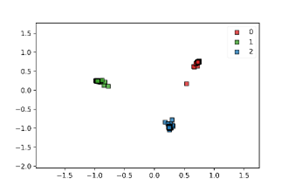}  \\
(a) & (b) \\
\includegraphics[width=0.5\columnwidth]{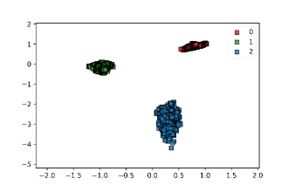} &
\includegraphics[width=0.5\columnwidth]{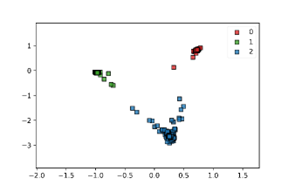}  \\
(c) & (d) \\
\end{tabular}
\caption{Learned feature representations of image samples: (a) the embeddings of the training samples returned by the proposed method trained with the balanced dataset, (b) the embeddings of the test samples returned by the proposed method trained with the balanced dataset, (c) the embeddings of the training samples returned by the proposed method trained with the imbalanced dataset, (d) the embeddings of the test samples returned by the proposed method trained with the imbalanced dataset.}
\label{fig:imbalanced}
\end{figure}

In our proposed method, the distances between the samples and their corresponding class centers are minimized independently of each other. Therefore, we expect that the proposed method will be more robust against to imbalanced datasets. To verify this, we conducted experiments on the same 3 Cifar-10 datasets used in illustrations experiments given in the paper.  We used the same deep neural network classifier yielding 2-dimensional feature spaces for this experiment. The number of training samples per class is 5000 for the selected classes and we first trained the proposed method by using the same amount of samples for each class. Then, we extracted the CNN features of test samples. After that, we decreased the number of samples of the blue colored classes to 500 (which is 10\% of the original size) to create an imbalanced training set. We trained another network by using this imbalanced dataset and extracted the CNN features of the testing samples. The visualization of the extracted features is shown in Fig. \ref{fig:imbalanced}, where the first row shows the CNN features of the training and test samples extracted by using the network trained wit the balanced dataset and the second row shows the extracted features by using the network trained wit imbalanced dataset. As seen in the figure, the extracted features of the test samples obtained by using the imbalanced dataset are similar to the one obtained by using the balanced dataset. This verifies that the proposed method is more robust against to imbalanced datasets as expected.

% or


%
%% use section* for acknowledgment
%\ifCLASSOPTIONcompsoc
%  % The Computer Society usually uses the plural form
%\else
%  % regular IEEE prefers the singular form
%\fi

\ifCLASSOPTIONcaptionsoff
  \newpage
\fi

\end{document}